\documentclass[letterpaper]{article} 
\usepackage{aaai25}  
\usepackage{times}  
\usepackage{helvet}  
\usepackage{courier}  
\usepackage[hyphens]{url}  
\usepackage{graphicx} 
\usepackage{natbib}  
\usepackage{caption} 
\usepackage{algorithm}
\usepackage{algorithmic}

\usepackage{subcaption}
\usepackage{newfloat}
\usepackage{listings}
\usepackage{color}

\usepackage{amstext}
\usepackage{amsmath}
\usepackage{amssymb}
\usepackage{multirow}
\usepackage{booktabs}
\usepackage{array}
\usepackage{subcaption}
\usepackage{cite}
\DeclareCaptionStyle{ruled}{labelfont=normalfont,labelsep=colon,strut=off} 
\floatstyle{ruled}
\newfloat{listing}{tb}{lst}{}
\floatname{listing}{Listing}
\title{Incomplete Multi-view Clustering via Diffusion Contrastive Generation}

\author{
    Yuanyang Zhang\textsuperscript{\rm 1}\equalcontrib,
    Yijie Lin\textsuperscript{\rm 2}\equalcontrib,
    Weiqing Yan\textsuperscript{\rm 3}, 
    Li Yao\textsuperscript{\rm 1,\rm 4}\thanks{Corresponding authors},
    Xinhang Wan\textsuperscript{\rm 5},\\
    Guangyuan Li\textsuperscript{\rm 6},
    Chao Zhang\textsuperscript{\rm 7},
    Guanzhou Ke\textsuperscript{\rm 8},
     Jie Xu\textsuperscript{\rm 9}
}
\affiliations{
     \textsuperscript{\rm 1}School of Computer Science and Engineering, Southeast University, Nanjing 210096, China\\
    \textsuperscript{\rm 2}College of Computer Science, Sichuan University\\
    \textsuperscript{\rm 3}School of Computer and Control Engineering, Yantai University\\
    \textsuperscript{\rm 4}Key Laboratory of New Generation Artificial Intelligence Technology and Its Interdisciplinary Applications (Southeast University), Ministry of Education, China\\
    \textsuperscript{\rm 5}College of Computer, National University of Defense Technology\\
    \textsuperscript{\rm 6}College of Computer Science and Technology, Zhejiang University\\
    \textsuperscript{\rm 7}Department of Control Science and Intelligence Engineering, Nanjing University \\
    \textsuperscript{\rm 8}School of Economics and Management, Beijing Jiaotong University\\
     \textsuperscript{\rm 9}School of Computer Science and Engineering, University of Electronic Science and Technology of China \\
     \{zhangyuanyang,~yao.li\}@seu.edu.cn,~linyijie.gm@gmail.com
}

\usepackage{bibentry}

\begin{document}

\maketitle

\begin{abstract}

Incomplete multi-view clustering (IMVC) has garnered increasing attention in recent years due to the common issue of missing data in multi-view datasets. The primary approach to address this challenge involves recovering the missing views before applying conventional multi-view clustering methods. Although imputation-based IMVC methods have achieved significant improvements, they still encounter notable limitations: 1) heavy reliance  on paired data for training the data recovery module, which is impractical in real scenarios with high missing data rates; 2) the generated data often lacks diversity and discriminability, resulting in suboptimal clustering results. 
To address these shortcomings, we propose a novel IMVC method called Diffusion Contrastive Generation (DCG). Motivated by the consistency between the diffusion and clustering processes, DCG learns the distribution characteristics to enhance clustering by applying forward diffusion and reverse denoising processes to intra-view data. By performing contrastive learning on a limited  set of paired multi-view samples, DCG can align the generated views with the real views, facilitating accurate recovery of views across arbitrary missing view scenarios. Additionally, DCG integrates instance-level and category-level interactive learning to exploit the consistent and complementary information available in multi-view data, achieving robust and end-to-end clustering. Extensive experiments demonstrate that our method outperforms state-of-the-art approaches. The code is available at https://github.com/zhangyuanyang21/2025-AAAI-DCG.

\end{abstract}


\section{Introduction}

In practical applications, data collected from different sensors or feature extraction methods is often presented in multiple-view forms. 
For instance, healthcare professionals utilize multi-view data such as MRI images, CT scans, and X-rays for diagnostic purposes~\cite{chen2022multi}. 
Similarly, autonomous vehicles employ data from cameras, LiDAR, and radar to perceive their surroundings and navigate~\cite{cui2024drive}. 
Multi-view data offers a comprehensive perspective of the same subject from various angles, making it a critical area of machine learning and data mining.

To extract and synthesize the full picture from these disparate sources, one of the pivotal techniques in multi-view learning is multi-view clustering (MVC), which aims to partition multi-view data into distinct clusters~\cite{cai2024dual,yan2024anchor,cai2024learning,lu2024survey} by leveraging the inherent consistency and complementarity nature of the information across different views~\cite{lin2024multi}. 
The success of existing multi-view clustering methods heavily relies on the completeness of multi-view data, i.e., all views are consistently available for each sample. However, this condition is rarely met in real-life scenarios due to sensor malfunctions or damage during storage.
To tackle this problem, Incomplete Multi-view Clustering (IMVC) has been proposed to effectively cluster data even when some views are missing~\cite{zhang2023robust,wan2024Fast}.

\begin{figure}[t]
    \centering
\includegraphics[width=\columnwidth]{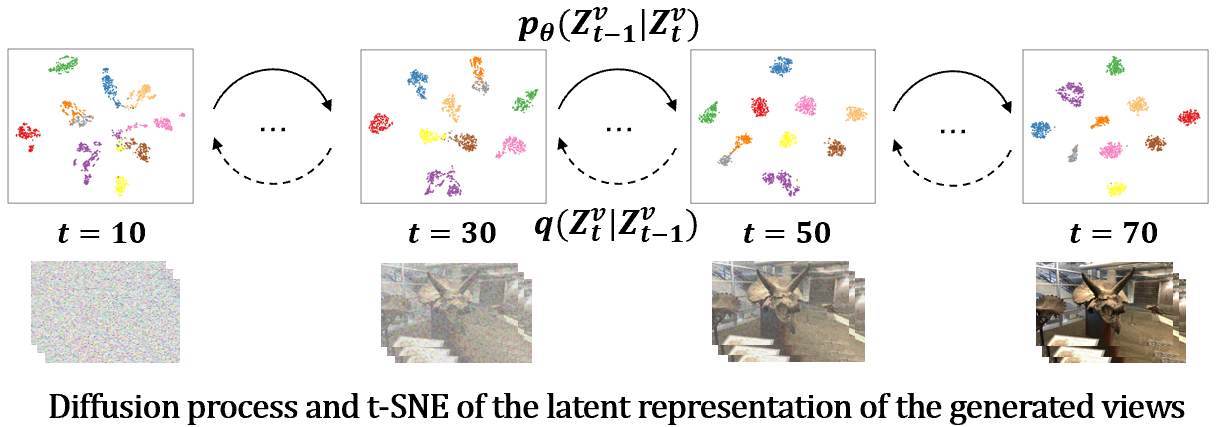}
    \caption{
    Motivation. We observed that as the diffusion process progresses, the generated sample representations tend to gradually aggregate toward the center of the cluster.
    }
\label{fig:observation}
\vspace{-2.5mm}
\end{figure}

The predominant strategy for addressing the incomplete multi-view data is to recover the missing view and subsequently apply existing MVC methods \cite{guan2024contrastive,ke2024rethinking,sun2024robust,cai2022superpixel} for clustering.
Specifically, existing IMVC methods primarily focus on crafting consistent common representations through contrastive learning and various data generation techniques to impute and fill in the missing data.
Among these, researchers have explored several imputation-based approaches, including cross-view neighbors-based methods~\cite{tang2022deep,jin2023deep}, cross-view prediction-based methods~\cite{lin2021completer,lin2022dual}, and generative adversarial network-based methods~\cite{wang2021generative}. Despite significant improvements achieved by these methods, they still encounter the following issues: 1) heavy reliance on paired data for training is difficult to satisfy in practical scenarios, especially in cases with high missing rates. For instance, meteorological monitoring stations are usually located in remote areas, making them vulnerable to weather and power supply issues, which results in high amounts of missing data; 2) the generated data often lacks diversity and discriminability, leading to suboptimal clustering results. For example, data recovery based on neighbor prediction may lack diversity, while that based on adversarial generation may lack discriminability due to the instability of model training.

Without necessitating adversarial training, diffusion models are renowned for their robust generative capabilities, presenting a promising alternative. For example,~\citet{wen2024} first incorporates the diffusion model into IMVC by using the available views as the condition for diffusion to generate the missing views.
However, this approach still heavily relies on paired data for diffusion completion and ignores the necessary connections between clustering and data completion.

To alleviate the above problems, we propose a novel incomplete multi-view clustering method termed diffusion contrastive generation (DCG), inspired by the consistency between the inverse diffusion process in data generation and the compactness of sample clusters. Our motivation is shown in Figure \ref{fig:observation}, as the diffusion process progresses, the generated sample representations gradually converge to the center of the category to which they belong. This phenomenon reveals the implicit connection between the diffusion process and data clustering, i.e., the inverse diffusion process not only facilitates data generation but also enhances the compactness of the data clusters.

Based on this phenomenon, DCG first applies forward diffusion and reverse denoising to the intra-view data to learn the distribution characteristics for learning clustering-friendly representations.
Then, by performing contrastive learning on only a small number of paired samples, we can align the generated views with the real views, enabling view recovery in arbitrary view missing scenarios. During the inference stage, by simply extrapolating the diffusion step (e.g., from 50 to 100), DCG obtains more desirable clustering results thanks to the diffusion and clustering consistency. Through DCG, we not only achieve view completion with only a small amount of paired data but also integrate the data completion and clustering processes into a unified diffusion process. Additionally, DCG integrates instance-level and category-level interactive learning to fully utilize consistent and complementary information in multi-view data, achieving end-to-end clustering.
The main contributions of this paper are summarized as follows:

\begin{itemize} 
\item 
We reveal that the diffusion process not only facilitates the recovery of missing views but also enhances the compactness of data clustering as the diffusion progresses. By extrapolating the diffusion step during the inference stage, we can obtain more compact clustering results thanks to the consistency between the diffusion process and clustering.
\item The proposed DCG method innovatively combines the diffusion process with contrastive learning to enable effective view generation and discriminative learning with limited paired data.
\item 
DCG achieves significant improvements across multiple datasets, especially in cases of high missing rates, where it outperforms state-of-the-art methods by 6.67\% on the CUB dataset with the missing rate of 0.7.
\end{itemize}
\begin{figure*}[!ht]
\centering
\includegraphics[width=1.0\linewidth]{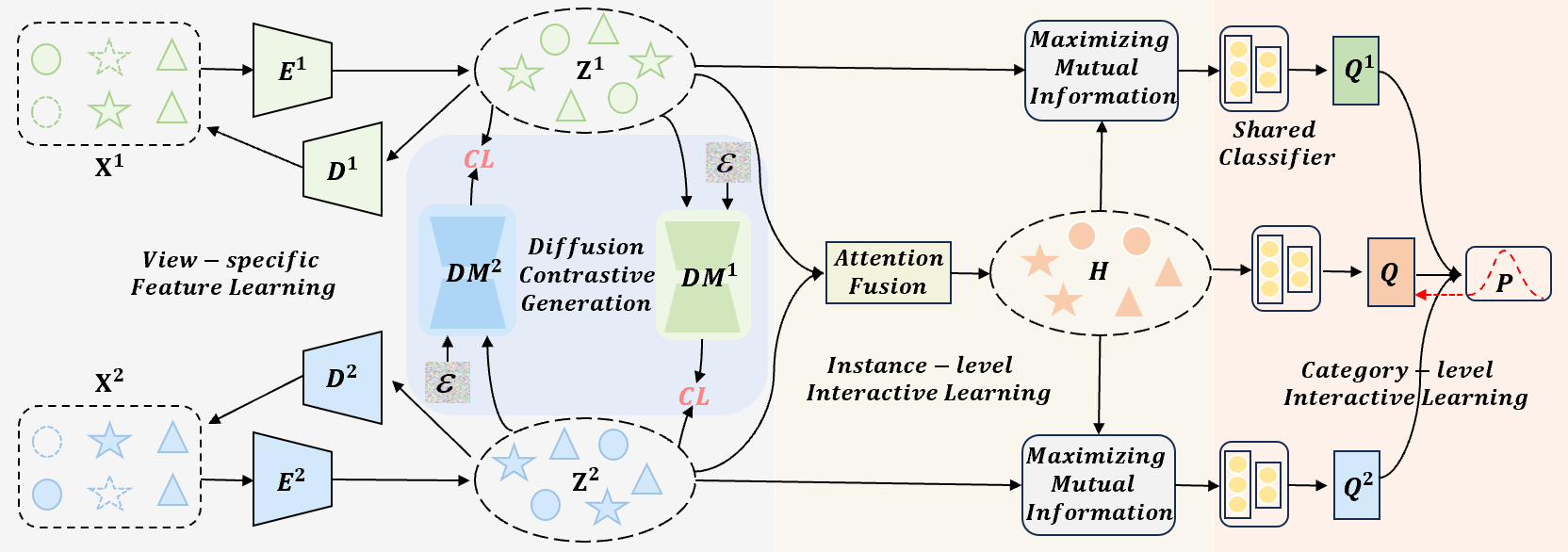}
\caption{
Without loss of generality, we take bi-view data as a showcase to demonstrate the overall framework of our proposed DCG.
As shown, our method is mainly divided into four modules: (1) View-specific Feature Learning; (2) Diffusion Contrastive Generation; (3) Instance-level Interactive Learning; (4) Category-level Interactive Learning. Note that, $\textit{E}$: Encoder; $\textit{D}$: Decoder; $\textit{DM}$: Diffusion Model; $\textit{CL}$: Contrastive Learning.}
\label{fig:framework}
\vspace{-2.5mm}
\end{figure*}

\section{Related Work} 
In this section, we briefly review recent advancements in two related topics, namely, deep incomplete multi-view clustering and diffusion model.

\subsection{Deep Incomplete Multi-view Clustering}

Inspired by the powerful feature representation capabilities of deep learning, many DIMVC methods have been developed. These methods can be categorized into two types: imputation-free methods and imputation-based methods. (1) Imputation-free methods aggregate the representations of existing views through different approaches and then obtain clustering results from the aggregated representations~\cite{xu2023adaptive,xu2024deep}.
(2) Imputation-based methods employ various strategies to impute missing views and subsequently perform clustering on the completed multi-view dataset. Specific methods include: i) cross-view neighbor-based approaches~\cite{tang2022deep,jin2023deep} that use cross-view nearest methods to impute missing data; ii) cross-view prediction-based methods~\cite{lin2021completer,lin2022dual} that maximize mutual information between different views of the samples and uses an encoder to estimate the missing views; iii) generative adversarial network-based methods~\cite{wang2021generative} that recover missing data through generation adversarial network.

\subsection{Diffusion Model} 

Diffusion Model~\cite{ho2020denoising} is a generative model that has achieved significant success in fields such as image generation~\cite{li2024self,xu2024ufogen}, image inpainting~\cite{lugmayr2022repaint}, and image super-resolution~\cite{li2024rethinking}. It utilizes a Markov diffusion process, gradually injecting Gaussian noise into the data, and then generates samples through a reverse denoising process. This approach has yielded impressive results in generating high-quality images. For instance, Li et al. utilize diffusion models to generate accurate prior knowledge, guiding the reconstruction of MRI images to maintain high quality without distortion~\cite{li2024rethinking}. Generative Adversarial Networks (GANs) have achieved success in incomplete multi-view clustering~\cite{wang2021generative} but face challenges such as training instability and model collapse~\cite{liu2023combating,luo2024dyngan}. In contrast, diffusion models based on Markov diffusion processes offer advantages of stable training processes and diverse generated results. Inspired by these studies, we propose the DCG approach, which aims to utilize the powerful generative capabilities of diffusion models to recover missing views from existing view information.

\section{Method}

\textbf{Notations}. Given a multi-view incomplete dataset $\{\mathbf{X}^{v} \in \mathbb{R}^{N_v \times D_v}\}_{v=1}^V$ which 
 consists of $N_v$ samples across $V$ views, where $\mathbf{X}^{v} = \{x_1^v, x_2^v, \ldots, x_{N_v}^v\} \in \mathbb{R}^{N_v \times D_v}$ denote the instance set of the $v$-th view, $D_v$ is the dimensionality of the samples, and $N_v$ denotes the number of samples. $K$ is the number of clusters.

\subsection{View-specific Feature Learning}

Deep autoencoder is widely used for unsupervised representation learning \cite{lin2021completer} by minimizing the reconstruction error. Considering that different views contain specific feature information, we construct a view-specific autoencoder to learn the latent representation \(\mathbf{Z}^v \in \mathbb{R}^{N_v \times d_v}\) by minimizing the reconstruction loss:
\begin{equation}
\label{eq:reconstruction loss}
\mathcal{L}_{\mathbf{R}}=\sum_{v=1}^{V}\sum_{i=1}^{N}\left\|\mathbf{X}_i^v-D_{\phi^v}^{(v)}\left(\mathbf{Z}_i^v\right)\right\|_2^{2},
\end{equation}
where $\mathbf{X}_i^v$ denotes the $i$-th sample of $\mathbf{X}^v$, and $D_{\phi^v}^{(v)}$ is the decoder of $v$-th view with the parameters $\phi^v$. The latent representation \(\mathbf{Z}_i^v\) is obtained as follows:
\begin{equation}
\mathbf{Z}_i^v = E_{\eta^{v}}^{(v)} \left(\mathbf{X}_i^v\right).
\end{equation}
where $E^{(v)}$ is the encoder of $v$-th view with parameters $\eta^{v}$.

\subsection{Diffusion Contrastive Generation}
To generate accurate missing view representations, inspired by the diffusion model~\cite{ho2020denoising} and contrastive learning~\cite{chen2020simple,lin2023graph}, we designed a Diffusion Contrastive Generation module.
Specifically, the module first applies forward diffusion and reverse denoising on the intra-view data to learn the distribution characteristics for clustering. Subsequently, by performing contrastive learning on a small number of paired samples, we can align the generated views with the real views, enabling view recovery in arbitrary view missing scenarios.
The Diffusion Contrastive Generation module includes the forward diffusion process, the reverse denoising process, and the data recovery process.

\noindent \textbf{Forward Diffusion Process.}
During the forward diffusion process, we gradually add Gaussian noise to the initial latent representation \( \mathbf{Z}^v_0 \) until it becomes fully Gaussian noise:
\begin{equation}
q\left(\mathbf{Z}^v_T \mid \mathbf{Z}^v_0\right)=\mathcal{N}\left(\mathbf{Z}^v_T ; \sqrt{\bar{\alpha}_T} \mathbf{Z}^v_0,\left(1-\bar{\alpha}_T\right) \mathbf{I}\right),
\end{equation}
where $T$ is the total number of time steps, $\mathcal{N}$ denotes the Gaussian distribution, $\alpha$ and $\bar{\alpha}_T$ are defined as:
$\alpha=1-\beta_t,\quad\bar{\alpha}_t=\prod_{i=1}^t \alpha_i$, where $t=(1,\dots,T)$, $\beta_{1:T} \in (0,1)$ are hyperparameters that control the variance of the noise.

\noindent \textbf{Reverse Denoising Process.}
The reverse denoising process is a Markov chain running backwards from $\mathbf{Z}_T^v$ to $\hat{\mathbf{Z}}^v$.
Taking the reverse step from $\mathbf{Z}^v_t$ to $\mathbf{Z}^v_{t-1}$ as an example:
\begin{equation}
p_{\theta}\left(\mathbf{Z}^v_{t-1} \mid \mathbf{Z}^v_t, \mathbf{Z}^v_0\right)=\mathcal{N}(\mathbf{Z}^v_{t-1} ; \boldsymbol{\mu}_t\left(\mathbf{Z}^v_t, \mathbf{Z}^v_0\right), \frac{1-\bar{\alpha}_{t-1}}{1-\bar{\alpha}_t} \beta_t \mathbf{I}), 
\end{equation}

\begin{equation}
\boldsymbol{\mu}_t\left(\mathbf{Z}^v_t, \mathbf{Z}^v_0\right)=\frac{1}{\sqrt{\alpha_t}}(\mathbf{Z}^v_t-\frac{1-\alpha_t}{\sqrt{1-\bar{\alpha}_t}} \boldsymbol{\epsilon}_t),
\end{equation}
where $\boldsymbol{\epsilon}_t$ represents the noise in $\mathbf{Z}^v_t$. Let $\boldsymbol{\epsilon}_\theta^v$ denotes the denoising network employed to estimate the noise $\boldsymbol{\epsilon}_t$ for view $v$, which is a U-Net architecture composed of multiple convolutional layers and ReLU activation functions. 
To encode the step information, we use sinusoidal and cosine positional encoding~\cite{vaswani2017attention} to encode the time step \( t \), namely,
\begin{equation}
\text{PE}(t) = \left[ \sin \left( \frac{t}{10000^{\frac{2i}{d}}} \right), \cos \left( \frac{t}{10000^{\frac{2i+1}{d}}} \right) \right],
\end{equation}
where \( d \) is the encoding dimension, and \( i = 0, 1, \ldots, \frac{d}{2}-1 \). The encoded time step vector is directly fed into the denoising network $\boldsymbol{\epsilon}_\theta^v$ to help the model understand the different stages of the generation process.

The objective of the diffusion model is to optimize the parameters \(\theta\) of the denoising network to make the estimated noise vector $\epsilon_\theta^v(\mathbf{Z}^v_t, \text{PE}(t))$ close to fully Gaussian noise \(\boldsymbol{\epsilon}\)~\cite{ho2020denoising}, namely,
\begin{equation} 
\mathcal{L}_{diff} = \sum_{v=1}^{V}\mathbb{E}_{\mathbf{Z}^v_0, t, \epsilon} \left[ \|\boldsymbol{\epsilon} - \boldsymbol{\epsilon}^v_\theta(\mathbf{Z}^v_t, \text{PE}(t))\|^2 \right].
\end{equation}

After the reverse denoising process, we achieved intra-view data generation. To achieve cross-view data generation for missing views through our diffusion model, we employ contrastive learning~\cite{chen2020simple} on a small number of paired samples to align the generated views with the real views. Specifically, we conduct the contrastive loss between the generated view $\hat{\mathbf{Z}}^m$ and all other existing views:
\begin{equation}
    \mathcal{L}_{gcl} = \frac{1}{2}\sum_{m=1}^V\sum_{n \neq m} \ell^{(mn)}_{cl},
    \label{eq:ins_loss}
\end{equation}
where the contrastive learning between generated view $\hat{\mathbf{Z}}^m$ and existing view $\mathbf{Z}^n$ is given as:
\begin{equation} 
\small
\ell^{(mn)}_{cl} = -\frac{1}{N}\sum_{i=1}^{N}\log \frac{e^{sim(\hat{\mathbf{Z}}_i^m, \mathbf{Z}_i^n)/\tau_F}}{\sum_{j=1}^{N}\sum_{v=m,n} e^{sim(\hat{\mathbf{Z}}_i^m, \mathbf{Z}_j^v)/\tau_F}},
\end{equation}

where $sim(\cdot, \cdot)$ denotes the 
cosine similarity, $\tau_F$ is the temperature parameter that controls the softness. 

To sum up, in the training stage, we employ $\mathcal{L}_{\mathbf{D}}$ to jointly train the entire module:
\begin{equation}
\mathcal{L}_{\mathbf{D}} = \mathcal{L}_{diff} + \mathcal{L}_{gcl}.
\end{equation}

\noindent\textbf{Data Recovery Process.} During the inference stage, we generate the missing \(i\)-th view representations from the other existing views, using the denoising network $\epsilon^i_\theta$ and iteratively applying the reverse denoising step \(p_\theta(\mathbf{Z}_{t-1}^v \mid \mathbf{Z}_{t}^v)\) from \(t=T_{ext}\) to \(t=1\) on each view. Formally,

\begin{equation} 
\bar{\mathbf{Z}}^j_{t-1} = \frac{1}{\sqrt{\alpha_t}} \left( \mathbf{Z}_t^j - \frac{1-\alpha_t}{\sqrt{1-\bar{\alpha}_t}}\boldsymbol{\epsilon}^{i}_\theta(\mathbf{Z}_t^j, \text{PE}(t))\right) + \sigma_t \xi,
\end{equation}
\begin{equation} 
\hat{\mathbf{Z}}^i= \frac{1}{V-1} \sum_{j \neq i}^V\bar{\mathbf{Z}}^j_0
\end{equation}
where \(\mathbf{Z}^j\) denotes the representations of the other existing views, \(\hat{\mathbf{Z}}^i\) denotes the generated missing representation of the \(i\)-th view, \(\sigma_t\) denotes the noise standard deviation associated with the time step \(t\), \(\xi \sim \mathcal{N}(0, \mathbf{I})\) denotes a noise term, and \(t=(T_{ext},\dots,1)\). 

\textbf{Remark.} Benefiting from the periodicity of the sinusoidal and cosine positional encoding~\cite{vaswani2017attention}, we can extrapolate the diffusion step during the inference stage (\textit{i.e.}, $T_{ext}> T$ ), leading to further clustering benefits.



\subsection{Instance-level Interactive Learning}

To fully exploit the cross-view consistency and complementarity of samples, we designed an instance-level interactive learning module that consists of two parts: an attention fusion module and a view alignment module. Directly concatenating or averaging these features usually leads to degraded clustering performance. To fully leverage complementary information from different views, we adopted an attention fusion module that automatically perceives fusion weights of views, which guide the modules in the network to reinforce each other~\cite{zhou2020end}. 
Specifically, the attention fusion module consists of three fully connected layers followed by a softmax layer. It takes the concatenated feature $\mathbf{Z}$ as input and produces a view weight vector $\mathbf{w}$ as output, namely,
\begin{equation}
\mathbf{Z} =\left[\mathbf{Z}^{1}, \mathbf{Z}^{2}, \ldots, \mathbf{Z}^{V}\right]
\end{equation}
\begin{equation}
\mathbf{w} = \text{Softmax}\left(\text{sigmoid}(\text{MLP}(\mathbf{Z}))
/\delta\right),
\end{equation}
where $\text{MLP}(\cdot)$ denotes three fully connected layers, and $\delta$ is a calibration factor. The Sigmoid function and calibration factor are used to prevent assigning scores close to one to the most relevant view features. 
Subsequently, the fused common representation \(\mathbf{H} \in \mathbb{R}^{N \times d_v}\) can be obtained by $\mathbf{H} = \sum_{v=1}^V \mathbf{w}_v\mathbf{Z}^{v}$.

Through the attention fusion module, the complementary information in multi-view data is fully utilized, which in turn guides the view-specific representation learning. Therefore, we maximize consistency between the common representation \( \mathbf{H} \) and the view-specific representation \(\mathbf{Z}^v\) by employing the following maximum mutual information loss~\cite{lin2022dual}, namely,
\begin{equation}\small
\begin{aligned}
\mathcal{L}_{\mathbf{I}} = & -\sum_{v=1}^{V} I\left(\mathbf{H} ; \mathbf{Z}^{v}\right) 
= -\sum_{v=1}^{V} \sum_{i,j} p\left(\mathbf{h}_{i}, \mathbf{z}_{j}^{v}\right) \log \frac{p\left(\mathbf{h}_{i}, \mathbf{z}_{j}^{v}\right)}{p\left(\mathbf{h}_{i}\right) p\left(\mathbf{z}_{j}^{v}\right)}.
\end{aligned}
\label{eq:MMI1}
\end{equation}
where \( I(\cdot) \) denotes the mutual information. By minimizing Eq.~(\ref{eq:MMI1}), the information correlation between the common representation and the view-specific representation among multiple views can be enhanced to effectively mine the cross-view instance consistency.

\subsection{Category-level Interactive Learning}
To obtain soft cluster assignments for end-to-end clustering, we design a shared classifier \( g(\cdot) \) with a softmax layer for all views to obtain the soft cluster assignments, i.e., $\mathbf{Q}^v=g(\mathbf{Z}^v)$, where $\mathbf{Q}^v \in \mathbb{R}^{N \times K}$. Due to the heterogeneous view-private information, and to improve the model's robustness, we adopt category-level contrastive loss ~\cite{chen2020simple} to achieve the cluster consistency. We define the category-level contrastive loss between $\mathbf{Q}^m$ and $\mathbf{Q}^n$ as follows:
\begin{equation} 
\small
\ell^{(mn)}_{sc} = -\frac{1}{K}\sum_{j=1}^K \log \frac{e^{sim(\mathbf{Q}^m_{.j},\mathbf{Q}^n_{.j})/\tau_C}}{\sum_{k=1}^K \sum_{v=m,n} e^{sim(\mathbf{Q}^m_{.j},\mathbf{Q}^v_{.k})/\tau_C}},
\end{equation}
where $\tau_C$ is the temperature parameter that controls the softness. For all views, the category-level contrastive loss can be denoted as:
\begin{equation} 
\mathcal{L}_{ccl} = \frac{1}{2}\sum_{m=1}^V \sum_{n \neq m} \ell^{(mn)}_{sc} + \sum_{v=1}^V \sum_{j=1}^K s^v_j \log s^v_j,
\end{equation}
\noindent where $s^v_j = \frac{1}{N}\sum_{i=1}^N q^v_{ij}$ is the regularization term, which is used to avoid all samples being assigned to the same cluster~\cite{huang2020deep}.

To further enhance the confidence of soft cluster assignments, we adopted KL divergence to guide complementary information learning through self-supervision, so that high-confidence instances not only improve their own representation learning, but also enhance the soft cluster assignments of other instances.
Specifically, we similarly input the common representation $\mathbf{H}$ into the shared classifier to obtain the corresponding soft cluster assignments, i.e., $\mathbf{Q}=g(\mathbf{H})$. By taking the maximum soft cluster assignment for each instance in $\mathbf{Q}$ and $\mathbf{Q}^v$, we can obtain high-confidence soft cluster assignment:
\begin{equation}
    \begin{split}
        \mathbf{q}_{ij} &= \mbox{max}\{\mathbf{Q}_{ij}, \mathbf{Q}_{ij}^v\}, \\
    \end{split}
    \label{eq:tar1}
\end{equation}
The high-confidence soft cluster assignment is used as the target assignment:
\begin{equation}
\begin{split}
    \mathbf{p}_{ij} &= \frac{\mathbf{q}_{ij}^2}{\sum_{j=1}^{k} \mathbf{q}_{ij}^2}, \\
\end{split}
\label{eq:tar2}
\end{equation}
Then, the clustering results are optimized by applying KL divergence to enhance the soft cluster assignment with high confidence and further blur the instances near the cluster boundaries:
\begin{equation}
    \mathcal{L}_{kl} = KL(\mathbf{P}\|\mathbf{Q}) = \sum_{i=1}^N \sum_{j=1}^K \mathbf{p}_{ij}\log\frac{\mathbf{p}_{ij}}{\mathbf{q}_{ij}},
    \label{eq:hg_loss}
\end{equation}
The overall loss of the category-level interactive learning module is:
\begin{equation}
    \mathcal{L}_{\mathbf{C}} =\mathcal{L}_{ccl} +  \mathcal{L}_{kl}.
    \label{eq:clil_loss}
\end{equation}
\subsection{The Objective Function}

Our model is an end-to-end clustering method that does not require k-means~\cite{bauckhage2015k} clustering to obtain the final clustering results. Therefore, we can optimize the entire model simultaneously. The total loss is as follows:
\begin{equation}
    \mathcal{L} = \mathcal{L}_{\mathbf{R}} + \lambda_1\mathcal{L}_{\mathbf{D}} + \lambda_2\mathcal{L}_{\mathbf{I}} + \lambda_3\mathcal{L}_{\mathbf{C}}
    \label{eq:overall_loss}.
\end{equation}
In our experiments, these three trade-off coefficients \(\lambda_1\), \(\lambda_2\), and \(\lambda_3\) are all set to \(1\).

\begin{table*}[!h]
\centering
\renewcommand{\arraystretch}{1.1}
\resizebox{\textwidth}{!}{
\begin{tabular}{|c|c|ccc|ccc|ccc|ccc|} 
\hline
\multirow{2}{*}{Dataset}     & Missing rates  & \multicolumn{3}{c|}{0.1}                      & \multicolumn{3}{c|}{0.3}                      & \multicolumn{3}{c|}{0.5}                      & \multicolumn{3}{c|}{0.7}                        \\ 
\cline{2-14}
                             & Evaluation metrics & ACC           & NMI           & ARI           & ACC           & NMI           & ARI           & ACC           & NMI           & ARI           & ACC           & NMI           & ARI            \\ 
\hline
\multirow{9}{*}{\rotatebox{90}{Synthetic3d}} 
                             & DCP~\cite{lin2022dual}   &88.00&65.17&67.87&79.83	&55.21	&56.69	&85.53	&58.33	&62.56	&81.50	&\underline{52.75}	&53.39\\
                             & DSIMVC ~\cite{tang2022deep}  & 73.11	&59.02	&58.38 &70.78	&56.41	&55.43	&67.44	&51.74	&49.42	&66.33	&48.96	&46.36\\
                             &GCFAgg ~\cite{yan2023gcfagg} &72.52&57.87&56.68&70.23&55.51&53.34&69.12&53.21&52.03&67.95&50.50&49.12\\
                             &CPSPAN\cite{jin2023deep} &\underline{88.83}&\underline{65.51}&\underline{69.12} &\underline{87.50} &\underline{61.79} &\underline{66.68}&\underline{86.33}&\underline{59.29}&\underline{63.65}&\underline{82.33}&52.24 &\underline{54.59}\\
                             &APADC~\cite{xu2023adaptive} &85.73	&59.07	&62.98	&86.47	&60.84	&64.90	&84.38	&58.48	&61.85	&80.83	&49.70	&53.42 \\
                             &ProImp~\cite{li2023incomplete}  &86.55	&61.79	&65.42	&85.50	&59.29	&62.81	&82.78	&54.76	&57.69	&76.67	&44.88	&46.42\\
                             &DVIMVC~\cite{xu2024deep}&50.03	&28.43	&24.57	&46.97	&22.07	&15.93	&50.32	&25.64	&18.79	&58.28	&36.84	&30.77  \\
                             &ICMVC~\cite{chao2024incomplete} &85.03	&58.30	&61.96	&87.17	&61.20	&66.29	&84.20	&55.33	&59.60	&70.77	&42.02	&40.71 \\
                             & DCG~(Ours) & \textbf{91.23}	&\textbf{71.35}	&\textbf{76.05}	&\textbf{88.00}&\textbf{64.11}	&\textbf{68.48}	&\textbf{87.67}&\textbf{63.23}	&\textbf{67.52}	&\textbf{85.50}	&\textbf{57.13}	&\textbf{62.17}\\ 
\hline
\multirow{10}{*}{\rotatebox{90}{CUB}} 
                            & DCP~\cite{lin2022dual} &42.77	&55.42	&33.39	&40.60	&52.37	&31.41	&38.87	&50.15	&31.18	&38.20	&47.97	&29.18\\ 
                            & DSIMVC ~\cite{tang2022deep} &63.67	&59.85	&46.23	&49.22	&49.93	&32.80	&49.89	&49.61	&32.35	&36.78	&40.30	&23.72\\
                            &GCFAgg ~\cite{yan2023gcfagg} &67.67&64.14&51.14&62.72&60.27&45.03&59.63&55.95&39.91&39.15&41.35&22.95\\
                            &CPSPAN\cite{jin2023deep} &\underline{76.67}&\underline{71.38}&\underline{58.65}&\underline{74.33} &\underline{70.33} &\underline{56.84}&\underline{73.33}&\underline{69.68}&\underline{57.17}&\underline{68.00}&\underline{68.36}&\underline{53.28}\\
                            &APADC~\cite{xu2023adaptive} &53.04	&59.05	&42.98	&51.43	&59.06	&40.52	&48.53	&59.93	&41.54	&42.57	&50.49	&31.91\\
                            &ProImp~\cite{li2023incomplete} &69.33	&69.65&55.93	&71.83	&64.45	&51.64	&69.56	&64.19	&52.06	&61.28	&56.37	&42.34\\
                            &DVIMVC~\cite{xu2024deep}&63.83	&63.00	&50.50	&70.42	&68.43	&56.35	&68.98	&64.81	&52.61	&55.53	&52.55	&39.54\\
                            &ICMVC~\cite{chao2024incomplete}  &32.33&33.36	&16.85	&45.13	&41.32	&25.11	&41.70	&37.78	&21.62	&39.30	&37.08	&21.06\\
                            & DCG~(Ours)) &\textbf{82.23}&\textbf{77.70}&\textbf{69.21}	&\textbf{77.17}	&\textbf{71.35}	&\textbf{59.85}	&\textbf{75.50}	&\textbf{72.21}	&\textbf{59.12}	&\textbf{74.67}	&\textbf{70.19}	&\textbf{56.41}\\
\hline
\multirow{9}{*}{\rotatebox{90}{HandWritten}}  
                            & DCP~\cite{lin2022dual}  &63.66 &70.44  &52.82 &75.68& 79.05&67.45 &72.07 &\underline{76.17}&63.81 &66.13 &69.04 &55.91\\ 
                            & DSIMVC \cite{tang2022deep} &71.37&70.24	&60.30	&67.47	&65.27	&54.26	&56.90	&57.99	&44.12	&49.27	&48.02	&34.10\\
                            &GCFAgg \cite{yan2023gcfagg} &75.07&68.19&59.79&69.16& 63.25&53.67 &61.02&53.57&42.15 &49.85&45.32&32.23\\
                            &CPSPAN\cite{jin2023deep} &68.70 &69.06 &59.17 &82.20 &75.05 &69.58 &72.55&68.60&57.51 &76.15 &69.04&\underline{61.33} \\
                            &APADC\cite{xu2023adaptive} &72.58	&71.29	&59.07	&60.66	&64.84	&48.75	&52.70	&63.12	&39.96	&53.68	&55.29	&36.20\\
                            &ProImp\cite{li2023incomplete}&81.05	&78.48	&70.40	&80.60	&77.26	&69.72	&\underline{78.98}	&74.08	&\underline{66.47}	&\underline{76.20}	&69.63	&61.10 \\
                            &DVIMVC\cite{xu2024deep} &29.08	&19.36	&11.06	&26.63	&16.57	&8.84	&25.13	&15.69	&8.09	&24.52	&14.25	&7.71\\
                            &ICMVC\cite{chao2024incomplete} &\underline{82.70} &\underline{81.06} &\underline{74.67} &\underline{82.27}	&\underline{79.90}	&\underline{72.81}& 75.04	&71.89	&63.29	&73.82	&\underline{70.15}	&60.76\\
                            & DCG (Ours) &\textbf{82.75} &\textbf{82.63} &\textbf{74.88} &\textbf{82.70}&\textbf{80.54}&\textbf{73.96}
                            & \textbf{80.80}& \textbf{76.21}& \textbf{70.45}
                            & \textbf{79.75}& \textbf{74.00}& \textbf{67.54}\\
\hline 
\multirow{9}{*}{\rotatebox{90}{LandUse-21}}  
                            & DCP~\cite{lin2022dual} &26.19&\underline{31.20}&13.02 &25.48&\underline{30.18}&11.08&21.52&26.32&11.11&22.18&27.00&10.13\\ 
                            & DSIMVC ~\cite{tang2022deep} &16.56	&16.40	&4.32	&16.46	&16.57	&4.35	&16.70	&16.84	&4.50	&16.22	&15.76	&4.12\\
                            &GCFAgg ~\cite{yan2023gcfagg}&19.05&19.99&6.32&18.53&19.89&6.24& 18.43& 19.47&5.98&19.05&20.33&6.37\\
                            &CPSPAN\cite{jin2023deep} &20.05 &27.20 &8.14 &18.05 &25.75 &6.90 &20.38 &26.99 &8.05 &23.52 &26.33 &10.31\\
                            &APADC~\cite{xu2023adaptive} & 20.92	&26.74	&7.97	&19.99	&24.40	&7.44	&19.15	&23.91	&7.08	&18.97	&21.44	&6.78\\
                                &ProImp~\cite{li2023incomplete} &24.43& 29.22&11.44 &24.63&28.45&11.78 &\underline{24.45}&\underline{27.43}&10.94 &\underline{23.86}&\underline{27.05}&\underline{10.82}\\
                            &DVIMVC~\cite{xu2024deep}&13.90	&22.48	&2.66	&13.24	&23.21	&2.74	&12.86	&19.67	&2.54	&12.57	&19.66	&2.57\\ 
                                &ICMVC~\cite{chao2024incomplete} &\underline{26.81}&30.71	&\underline{13.60}	&\underline{26.12}	&29.83	&\underline{12.87}	&24.43	&27.27	&\underline{11.12}	&22.97	&25.15	&9.65\\
                            & DCG~(Ours) &\textbf{27.52}&\textbf{31.36}&\textbf{14.57}&\textbf{27.33}&\textbf{32.09}&\textbf{14.47}&\textbf{25.76}&\textbf{29.07}&\textbf{13.17}&\textbf{25.14}&\textbf{27.23}&\textbf{11.85}\\
                            \hline
\multirow{10}{*}{\rotatebox{90}{Fashion}}  
                            & DCP~\cite{lin2022dual} &83.70	&84.30	&76.50	&71.80	&70.90	&52.50	&60.80	&59.50	&33.10	&49.90	&48.40	&19.10\\ 
                            & DSIMVC ~\cite{tang2022deep} &88.00	&86.40	&81.10	&87.30	&85.00	&78.90	&83.50	&80.30	&73.70	&75.71&71.53	&\underline{69.00}\\
                            &GCFAgg ~\cite{yan2023gcfagg} &78.21&74.50 &66.28 &76.34&72.53&63.98 &74.47&69.83&60.37 &72.47&67.93&57.98 \\
                            &CPSPAN\cite{jin2023deep} &66.16 &68.45 &55.73 &64.80 &68.22 &55.55 &54.81 &64.13 &48.84 &66.32 &68.44 &55.99\\
                            &APADC~\cite{xu2023adaptive} &81.40	&86.50	&73.30	&80.90	&\underline{85.01}	&73.10	&75.40	&81.50	&67.60	&52.90	&59.78	&37.40\\
                            &ProImp~\cite{li2023incomplete} &\underline{92.88}	&\underline{88.34}	&\underline{86.09}	&74.11&	76.97	&66.42	&\underline{89.76}	&\underline{81.94}	&\underline{79.93}	&76.28	&\underline{74.44}	&66.71\\
                            &DVIMVC~\cite{xu2024deep} &79.38& 80.22& 71.50& 82.26& 79.82&72.46& 80.17&77.09 &69.51 &\underline{76.95}& 74.25 &65.52\\ 
                            &ICMVC~\cite{chao2024incomplete} &92.41	&87.05	&85.11	&\underline{89.31}	&83.06	&\underline{79.93}	&79.37	&74.44	&68.46	&72.17	&68.45	&59.88\\
                            & DCG~(Ours)&\textbf{95.83} &\textbf{91.29} &\textbf{91.19} &\textbf{93.13} &\textbf{86.99} &\textbf{86.00} &\textbf{90.04} &\textbf{82.25} &\textbf{79.99} 	&\textbf{85.76} &\textbf{76.42} &\textbf{72.79} \\
\hline
\end{tabular}}
\caption{Clustering results of all methods on five datasets with different missing rates. The best and second-best results are highlighted in bold and underlined, respectively.}
\label{tab:incomplete performance}
\end{table*}

\begin{table*}[h]
\centering
\renewcommand{\arraystretch}{1.1}
\resizebox{\textwidth}{!}{
\begin{tabular}{l|ccccc|ccc|ccc|ccc}
\hline
\multicolumn{1}{c|}{\multirow{2}{*}{Variants}} & \multicolumn{5}{c|}{Components} & \multicolumn{3}{c|}{CUB} & \multicolumn{3}{c|}{HandWritten} & \multicolumn{3}{c}{Fashion} \\
\cline{2-15}
 & $\mathcal{L}_{diff}$ & $\mathcal{L}_{gcl}$ & $\mathcal{L}_{\mathbf{I}}$ & $\mathcal{L}_{ccl}$ & $\mathcal{L}_{kl}$ & ACC & NMI & ARI & ACC & NMI & ARI & ACC & NMI & ARI \\
\hline
(\emph{w/o}) $diff$ & $\times$ & $\checkmark$ & $\checkmark$ & $\checkmark$ & $\checkmark$ & 70.96 & 67.35 & 53.49 & 74.65 & 74.13 & 63.44 & 81.69 & 80.68 & 72.57 \\
(\emph{w/o}) $gcl$ & $\checkmark$ & $\times$ & $\checkmark$ & $\checkmark$ & $\checkmark$ & 71.46 & 67.38 & 53.87 & 75.95 & 75.36 & 65.99 & 83.41 & 81.10 & 75.63 \\
(\emph{w/o}) $diff$\&$gcl$ & $\times$ & $\times$ & $\checkmark$ & $\checkmark$ & $\checkmark$ & 60.73 & 61.21 & 42.23 & 66.00 & 66.41 & 54.82 & 75.62 & 73.10 & 64.65 \\
(\emph{w/o}) $\mathbf{I}$ & $\checkmark$ & $\checkmark$ & $\times$ & $\checkmark$ & $\checkmark$ & 72.83 & 68.25 & 56.78 & 77.11 & 74.65 & 66.78 & 85.59 & 84.21 & 77.56 \\
(\emph{w/o}) $ccl$ & $\checkmark$ & $\checkmark$ & $\checkmark$ & $\times$ & $\checkmark$ & 71.23 & 66.78 & 53.78 & 73.72 & 72.65 & 62.78 & 80.56 & 78.33 & 72.89 \\
(\emph{w/o}) $kl$ & $\checkmark$ & $\checkmark$ & $\checkmark$ & $\checkmark$ & $\times$ & 73.63 & 67.98 & 56.53 & 79.23 & 77.82 & 71.63 & 87.80 & 83.22 & 79.46 \\
(\emph{w/o}) $ccl$\&$kl$ & $\checkmark$ & $\checkmark$ & $\checkmark$ & $\times$ & $\times$ & 68.89 & 65.34 & 51.29 & 70.79 & 70.23 & 61.66 & 78.57 & 76.58 & 70.28 \\
Full model & $\checkmark$ & $\checkmark$ & $\checkmark$ & $\checkmark$ & $\checkmark$ & \textbf{77.17} & \textbf{71.35} & \textbf{59.85} & \textbf{82.70} & \textbf{80.54} & \textbf{73.96} & \textbf{93.13} & \textbf{86.99} & \textbf{86.00} \\
\hline
\end{tabular}
}
\caption{Ablation study on CUB, HandWritten, and Fashion with the missing rate of 0.3.}
\label{tab:ablation}
\end{table*}

\begin{table}[h]
\centering
\renewcommand{\arraystretch}{1.1}
\setlength{\tabcolsep}{8pt} 
\begin{tabular}{lccc}
\hline
Strategies    & ACC     & NMI     & ARI     \\ \hline
GAN~\cite{wang2021generative}           & 55.82  & 53.43  & 48.39  \\
Prediction~\cite{lin2022dual}   & 71.13  & 66.76  & 54.61  \\
Paired        & 72.83  & 66.80  & 53.98  \\ \hline
Default       & \textbf{77.17} & \textbf{71.35} & \textbf{59.85} \\ \hline
\end{tabular}
\caption{Ablation study on different data recovery and data usage strategies on CUB with the missing rate of 0.3. ``Paired'' denotes use only paired data.}
\label{tab:ablation2}
\end{table}

\begin{figure*}[h]
    \centering
    \subfloat[ACC vs. $\lambda1$]{
       \includegraphics[width=0.24\linewidth]{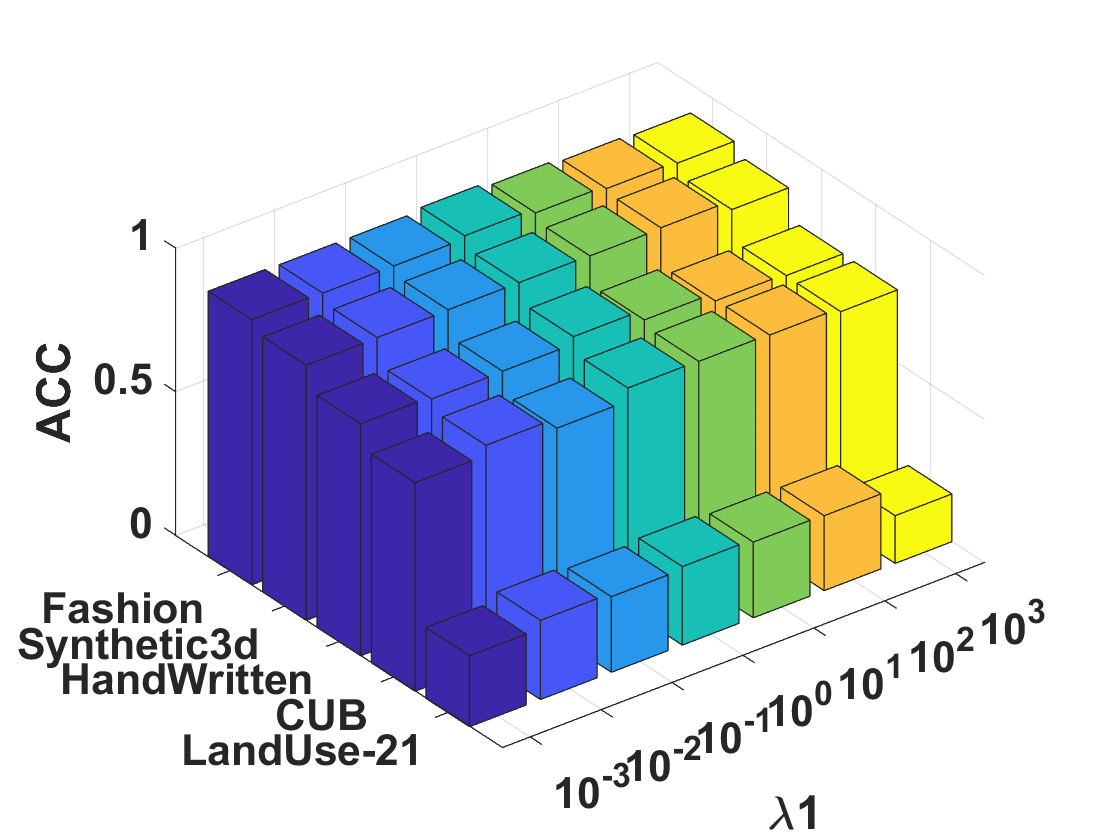}}
                 \subfloat[ACC vs. $\lambda2$]{
        \includegraphics[width=0.24\linewidth]{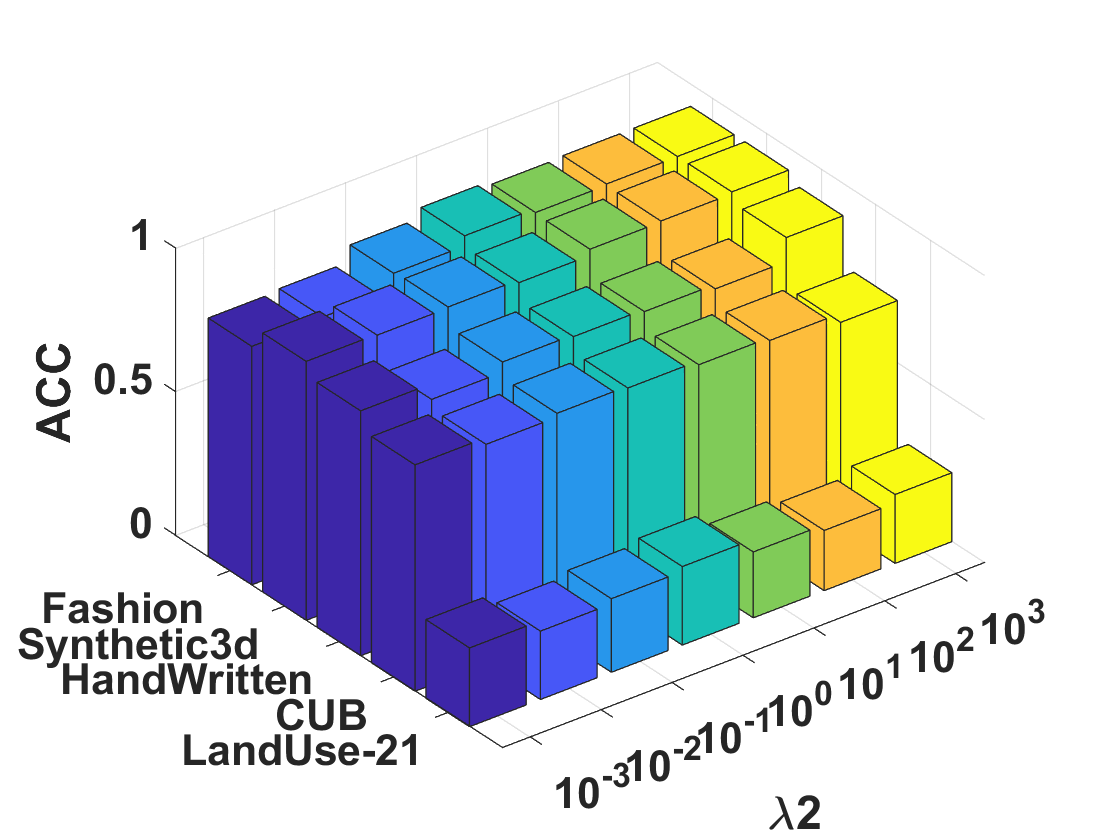}}
        \subfloat[ACC vs. $\lambda3$]{
        \includegraphics[width=0.24\linewidth]{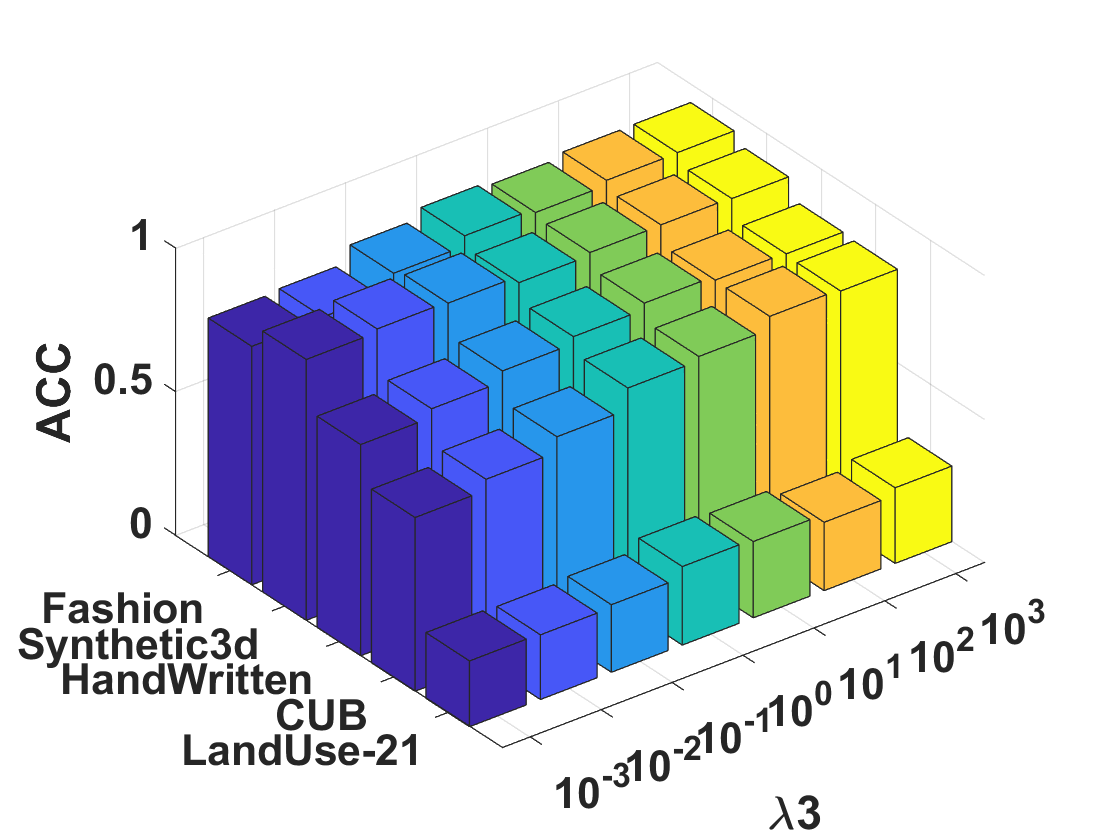}}
        \subfloat[ACC vs. $T$]{
        \includegraphics[width=0.24\linewidth]{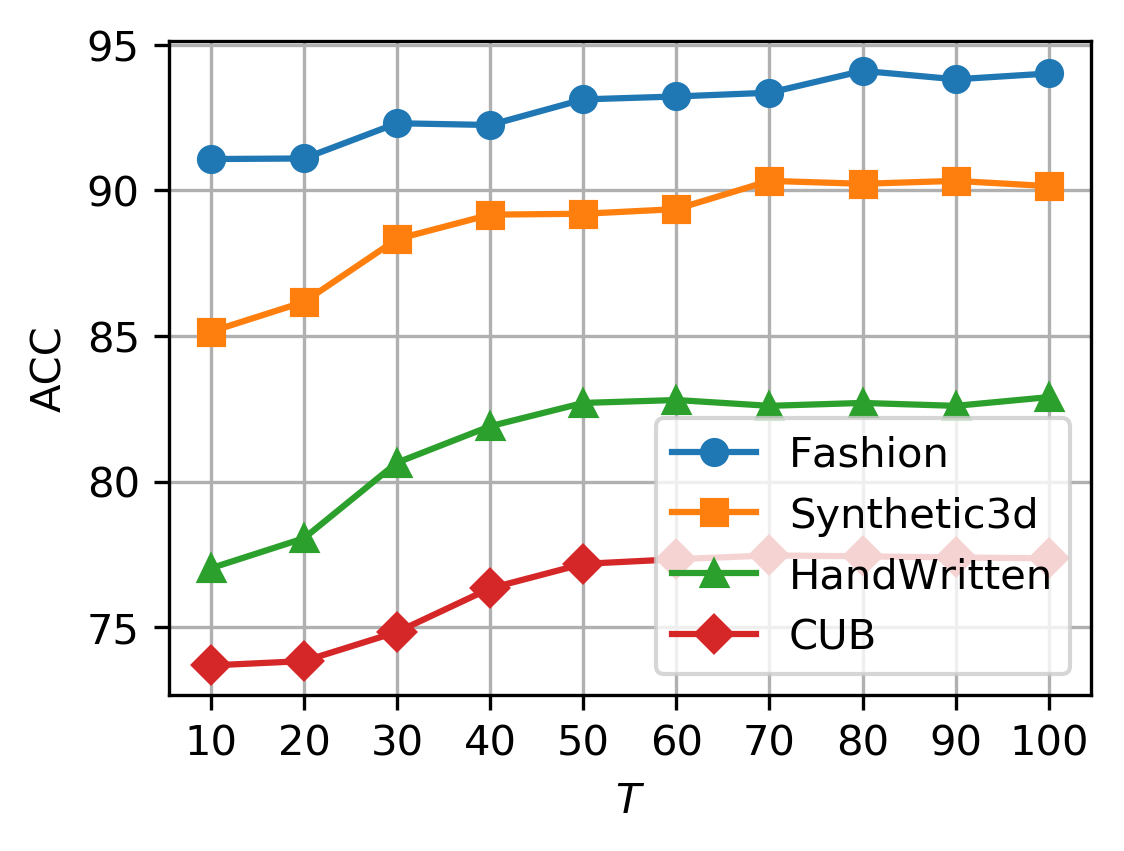}}
    \caption{Parameter sensitivity analysis on different datasets with the missing rate of 0.3.}
    \label{fig:psa} 
    \vspace{-1.5mm}
\end{figure*}

\begin{figure}[h]
    \centering
        \subfloat[Raw Features]{
        \includegraphics[width=0.48\linewidth]{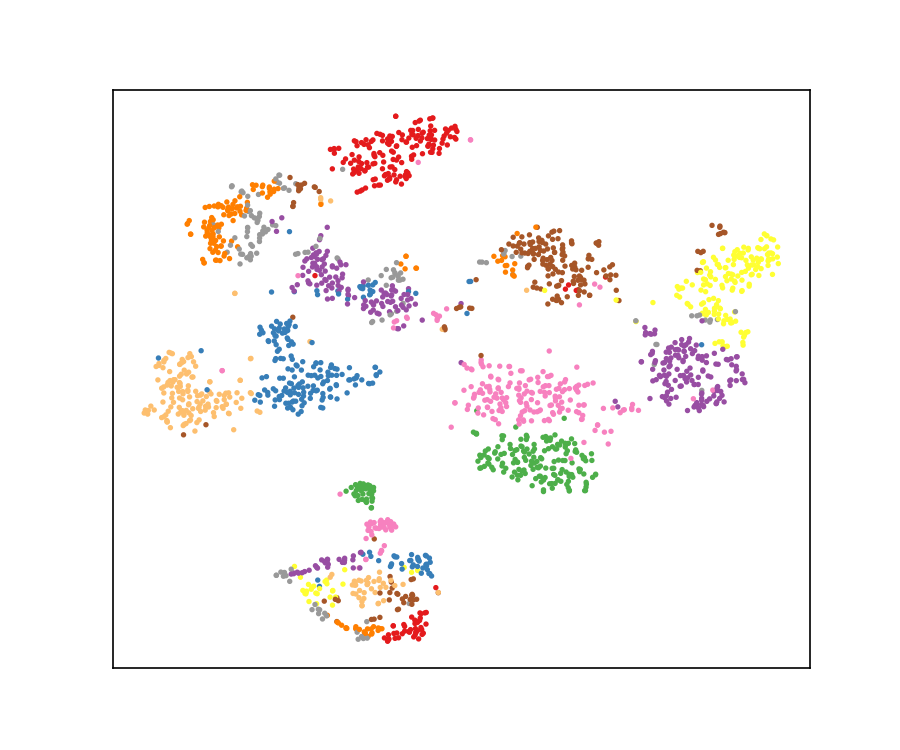}}
        \subfloat[Final Result]{
        \includegraphics[width=0.48\linewidth]{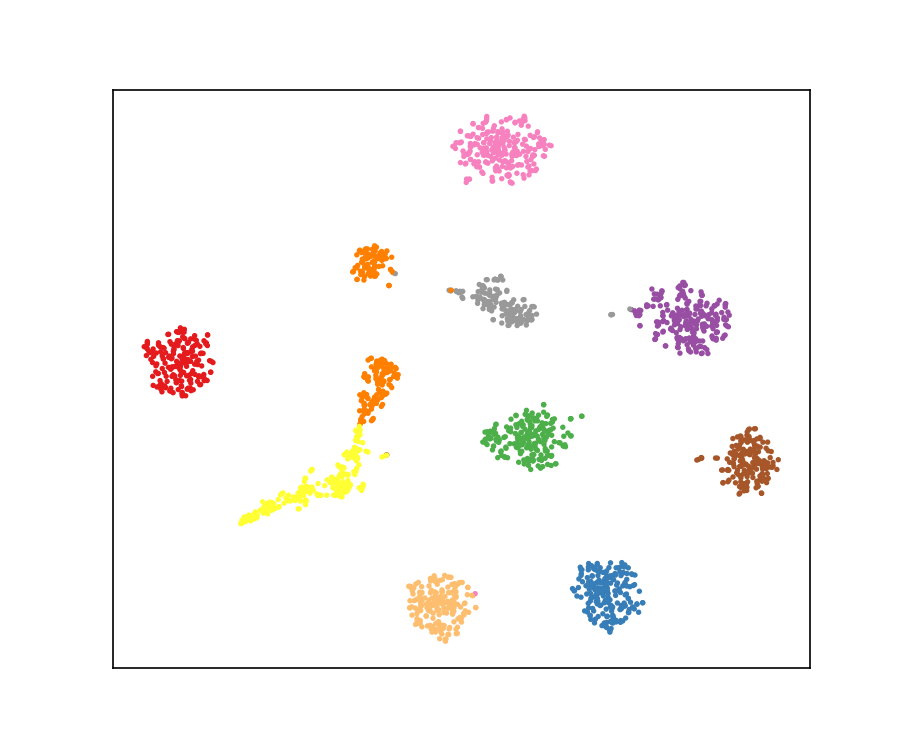}}
	  \caption{t-SNE visualization of HandWritten with the missing rate of 0.3.}
	  \label{fig:TSE}
\vspace{-2.5mm}
\end{figure}

\section{Experiments}

To evaluate the effectiveness of DCG, we conducted extensive experiments to answer the following questions:
(Q1) Does DCG outperform state-of-the-art IMVC methods?
(Q2) Does each component of DCG contribute to the overall performance?
(Q3) Do the hyperparameters affect the performance of DCG?
(Q4) What is the clustering structure revealed by DCG?

\subsection{Experimental Setup}

We conducted experiments on five multi-view datasets, including Synthetic3D~\cite{kumar2011co}, CUB~\cite{wah2011caltech}, HandWritten~\cite{asuncion2007uci}, LandUse-21~\cite{yang2010bag}, and Fashion~\cite{xiao2017fashion}. To evaluate the performance of handling incomplete multi-view data, following \cite{lin2021completer}, we randomly select \( m \) instances and randomly delete one view, where the missing rate is \( m/n \), with \( n \) representing the total number of instances. For a comprehensive analysis, three widely-used clustering metrics, including Accuracy (ACC), Normalized Mutual Information (NMI), and Adjusted Rand Index (ARI), are used. A higher value of these metrics indicates better clustering performance. Detailed descriptions of the datasets and implementation are in the supplementary materials.




\subsection{Comparisons with State of the Arts (Q1)}

We compared DCG with eight state-of-the-art IMVC methods, including DCP~\cite{lin2022dual}, DSIMVC~\cite{tang2022deep}, GCFAgg~\cite{yan2023gcfagg}, CPSPAN~\cite{jin2023deep}, APADC~\cite{xu2023adaptive}, ProImp~\cite{li2023incomplete}, DVIMVC~\cite{xu2024deep}, and ICMVC~\cite{chao2024incomplete}. A detailed description of these baselines is provided in the supplementary materials. 

We evaluated DCG and baselines with different missing rates.
Table~\ref{tab:incomplete performance} reports the average clustering performance under five random experiments. It can be observed that:
(1) Our DCG method outperforms all baselines methods in all cases. In particular, when the missing rate is 0.1 on CUB, DCG achieves 5.56\%, 6.32\%, and 10.56\% improvements on ACC, NMI, and ARI, respectively, compared to the second best method. These results demonstrate the effectiveness and superiority of our proposed method over the baselines. 
(2) As the missing rates increase, the performance of all methods generally declines, indicating the missing data hinders the sufficient exploration of multi-view data. However, compared with other approaches, our DCG demonstrates greater robustness against increasing missing rates in most cases. For example, on HandWritten, when the missing rate increase from 0.1 to 0.7, the accuracy drop of DCG is 3\%, while that of ICMVC is 8.88\%, which demonstrates the stability of DCG.

\subsection{Ablation Studies (Q2)}

To evaluate the effectiveness of each component in our proposed DCG method, we conducted ablation experiments on CUB, HandWritten, and Fashion. Specifically, we designed several variants: ``(\emph{w/o}) $diff$", ``(\emph{w/o}) $gcl$", and ``(\emph{w/o}) $diff$\&$gcl$," which respectively represent the removal of the diffusion model, the removal of the contrastive learning loss, and the removal of both. Moreover, ``(\emph{w/o}) $\mathbf{I}$" indicates the removal of the maximum mutual information loss. Similarly, ``(\emph{w/o}) $ccl$", ``(\emph{w/o}) $kl$", and ``(\emph{w/o}) $ccl$\&$kl$" respectively represent the removal of the category-level contrastive loss, the removal of the KL divergence loss, and the removal of both.

As shown in Table \ref{tab:ablation}, we can draw the following conclusions: 1) The model performance decreases when any module in the DCG is removed, indicating that each module contributes significantly to improving the overall performance; 2) Removing both the diffusion model and the contrastive learning loss simultaneously results in the largest decrease in model performance, suggesting that our diffusion contrastive generation module significantly contributes to recovering missing view representations.

To further evaluate the effectiveness of the diffusion contrastive generation module, we replaced it with GAN~\cite{wang2021generative} and Prediction~\cite{lin2022dual}, as shown in Table \ref{tab:ablation2}. The results show that our recovery strategy outperforms other methods, primarily because it innovatively combines diffusion processes with contrastive learning, enabling effective view generation and discriminative learning. Additionally, to verify the effectiveness of our data usage strategy, we changed the strategy to use only paired data for experiments. As shown in Table \ref{tab:ablation2}, the results show that, compared to the strategy of using only paired data, our data usage strategy can more effectively utilize the multi-view data, thereby improving clustering performance.

\subsection{Hyperparameters Analysis (Q3)}

To evaluate the robustness of our proposed method DCG to hyperparameters, we conducted experiments on different datasets with missing rate of 0.3 to analyze the impact of the three trade-off coefficients $\lambda_1$, $\lambda_2$, and $\lambda_3$, and the total number of time steps $T$ on the clustering performance of DCG. Figure \ref{fig:psa} shows the ACC of our method DCG as the trade-off coefficients vary from $10^{-3}$ to $10^{3}$ and the total number of time steps $T$ varies from $10$ to $100$. The results indicate that DCG's clustering performance is insensitive to changes in $\lambda_1$, $\lambda_2$, and $\lambda_3$ within the range of $10^{-1}$ to $10^{1}$. Additionally, the ACC gradually increases by extrapolating the diffusion step from 50 to 100, which validates our findings on the consistency between diffusion processes and data clustering.

\subsection{Visualization Analysis (Q4)}

To intuitively demonstrate the superiority of DCG, we use t-SNE to visualize both the raw features and the latent representations learned by DCG, as shown in Figure \ref{fig:TSE}. We observe that after training our model, the instances from the same cluster become more compact, and the instances of different clusters are separated further apart. This indicates that our DCG method can effectively leverage the consistency and complementarity of information in multi-view data.

\section{Conclusion}

In this paper, we find that the diffusion process not only helps recover missing views but also enhances the compactness of data clustering as the diffusion progresses. Based on this phenomenon, we propose a novel Diffusion Contrastive Generation (DCG) method, which innovatively combines the diffusion process with contrastive learning, thereby enabling effective view generation and discriminative learning using limited paired data. Moreover, to further improve clustering performance, DCG integrates instance-level and category-level interactive learning to fully exploit the consistent and complementary information in multi-view data, achieving end-to-end clustering. Extensive experimental results demonstrate the effectiveness and superiority of our method on the IMVC task.

\section{Acknowledgments}
This research work is supported by the Big Data Computing Center of Southeast University.

\bibliography{aaai25}

\end{document}